\documentclass[lettersize,journal]{IEEEtran}
\usepackage{amsmath,amssymb,amsfonts}
\usepackage{algorithmic}
\usepackage{algorithm}
\usepackage{array}

\usepackage{textcomp}
\usepackage{stfloats}
\usepackage{url}
\usepackage{verbatim}
\usepackage{graphicx}
\usepackage{cite}
\usepackage{textcomp}
\usepackage{graphics} % for pdf, bitmapped graphics files
\usepackage{epsfig} % for postscript graphics files
\usepackage{color}
\usepackage{bm}
\usepackage{subcaption}
\usepackage{multirow}
\usepackage{array}
\usepackage{booktabs}

\usepackage{mathtools}
\usepackage{cite}
\usepackage{cuted}
\definecolor{pointcolor}{RGB}{0,114,189}
\definecolor{linecolor}{RGB}{217,83,25}
\definecolor{planecolor}{RGB}{237,177,32}
\definecolor{spherecolor}{RGB}{119,172,48}
\definecolor{ellipsoidcolor}{RGB}{126,47,142}
\definecolor{cylindercolor}{RGB}{77,190,238}
\definecolor{conecolor}{RGB}{162,20,47}

\DeclareMathOperator*{\argmin}{arg\,min}

\graphicspath{{figures/}}
\begin{document}

\title{2D-3D Pose Tracking with Multi-View Constraints}

\author{Huai Yu$^*$, \IEEEmembership{Member, IEEE}, Kuangyi Chen$^*$, Wen Yang, \IEEEmembership{Senior Member, IEEE},  Sebastian Scherer, \IEEEmembership{Senior Member, IEEE}, and Gui-Song Xia, \IEEEmembership{Senior Member, IEEE}
\thanks{$^*$These authors contribute equally to this work. }
\thanks{This work has been submitted to the IEEE for possible publication. Copyright may be transferred without notice, after which this version may no longer be accessible.}
%Manuscript received x, 2023; revised x, 2023; accepted x, 2023; The work was partially supported by the National Natural Science Foundation of China under Grant 62301370, the Natural Science Foundation of Hubei Province, China under Grant 2022CFB600, the GuangDong Basic and Applied Basic Research Foundation under Grant 2022A1515110032.}
\thanks{Huai Yu, Kuangyi Chen and Wen Yang are with the School of Electronic Information, Wuhan University, Wuhan 430072, China (e-mail: yuhuai@whu.edu.cn, chenky721@whu.edu.cn, yangwen@whu.edu.cn)}
% \thanks{Huai Yu is also with the Wuhan Unversity Shenzhen Research Institute, Shenzhen 518057, China.}
\thanks{Sebastian Scherer is with the Robotics Institute, Carnegie Mellon University, Pittsburgh, PA 15213, USA (e-mail: basti@andrew.cmu.edu)}
\thanks{Gui-Song Xia is with the School of Computer Science, Wuhan University, Wuhan 430072, China (e-mail: guisong.xia@whu.edu.cn)}
}

\markboth{Journal of \LaTeX\ Class Files,~Vol.~14, No.~8, August~2021}%
{Shell \MakeLowercase{\textit{et al.}}: A Sample Article Using IEEEtran.cls for IEEE Journals}

\IEEEpubid{0000--0000/00\$00.00~\copyright~2021 IEEE}

\maketitle

\begin{abstract}
% Camera localization in 3D LiDAR maps achieves localization by establishing 2D-3D correspondences between 2D camera frames and 3D LiDAR points. However, existing methods mostly focus on solving the cross-modal gaps without considering the multi-view constraints in the learning process, leading to the estimated poses that jitter over continuous frames. To address this issue, we propose a novel framework that tightly couples 2D-3D and 2D-2D correspondences during the learning and inference process. Firstly, we fuse two image-to-depth flow estimation networks and one optical flow estimation network to simultaneously estimate the 2D-3D and 2D-2D correspondences between adjacent camera frames and LiDAR points. More importantly, we devise a cross-modal consistency loss for network supervision, which enables the localization model to learn using the multi-view constraints. Then in the inference process, we design an objective function for nonlinear optimization of the estimated camera poses corresponding to adjacent camera frames. Extensive experiments on the KITTI dataset demonstrate that the proposed method can smooth and improve the 2D-3D pose tracking performance.
%camera pose tracking 
% Camera localization in 3D LiDAR maps has gained increasing attention due to its balance between cost and accuracy.
Camera localization in 3D LiDAR maps has gained increasing attention due to its promising ability to handle complex scenarios, surpassing the limitations of visual-only localization methods.
However, existing methods mostly focus on addressing the cross-modal gaps, estimating camera poses frame by frame without considering the relationship between adjacent frames, which makes the pose tracking unstable. To alleviate this, we propose to couple the 2D-3D correspondences between adjacent frames using the 2D-2D feature matching, establishing the multi-view geometrical constraints for simultaneously estimating multiple camera poses. Specifically, we propose a new 2D-3D pose tracking framework, which consists: a front-end hybrid flow estimation network for consecutive frames and a back-end pose optimization module. We further design a cross-modal consistency-based loss to incorporate the multi-view constraints during the training and inference process. We evaluate our proposed framework on the KITTI and Argoverse datasets. Experimental results demonstrate its superior performance compared to existing frame-by-frame 2D-3D pose tracking methods and state-of-the-art vision-only pose tracking algorithms. More online pose tracking videos are available at \url{https://youtu.be/yfBRdg7gw5M}.
\end{abstract}

\section{Introduction}
\IEEEPARstart{C}{amera} localization in 3D LiDAR maps has attracted more and more attention due to the convenient system setup and high application prospects for mobile robots and autonomous driving. On the one hand, it only leverages lightweight and low-cost cameras, similar to visual SLAM, while offering significant potential in mitigating pose drift issues. On the other hand, LiDAR maps can be effortlessly constructed at a large scale and remain unaffected by changes in illumination, thanks to LiDAR SLAM or registration techniques. However, the existence of cross-modal gaps between camera images and LiDAR point clouds hampers the establishment of robust 2D-3D correspondences, consequently challenging the robustness and accuracy of camera localization.

% Extensive research focuses on solving the cross-modal gaps in between no matter whether using traditional geometric or learning-based strategies. Utilizing the acquired 2D-3D correspondences between camera images and point clouds, the six-degree-of-freedom (6-DoF) pose of the camera in the LiDAR map can be readily computed. However, this frame-by-frame pose estimation approach neglects the multi-view constraints between consecutive video frames, resulting in an unstable pose tracking process.

% However, the effective integration of deep learning techniques to tightly couple 2D-3D and 2D-2D correspondences remains an open problem.
% On the one hand, it uses lightweight and low-cost cameras same to visual SLAM but has great potential to overcome the pose drift issues. On the other hand, LiDAR maps can be easily built on large-scale and illumination-invariant using LiDAR SLAM or registration techniques. Extensive research focuses on solving the cross-modal gaps in between no matter whether using traditional geometric or learning-based strategies. However, the tight couple of 2D-3D and 2D-2D correspondences using deep learning still remains an open problem. 

\begin{figure}[t]
    \centering
    \includegraphics[width=\linewidth]{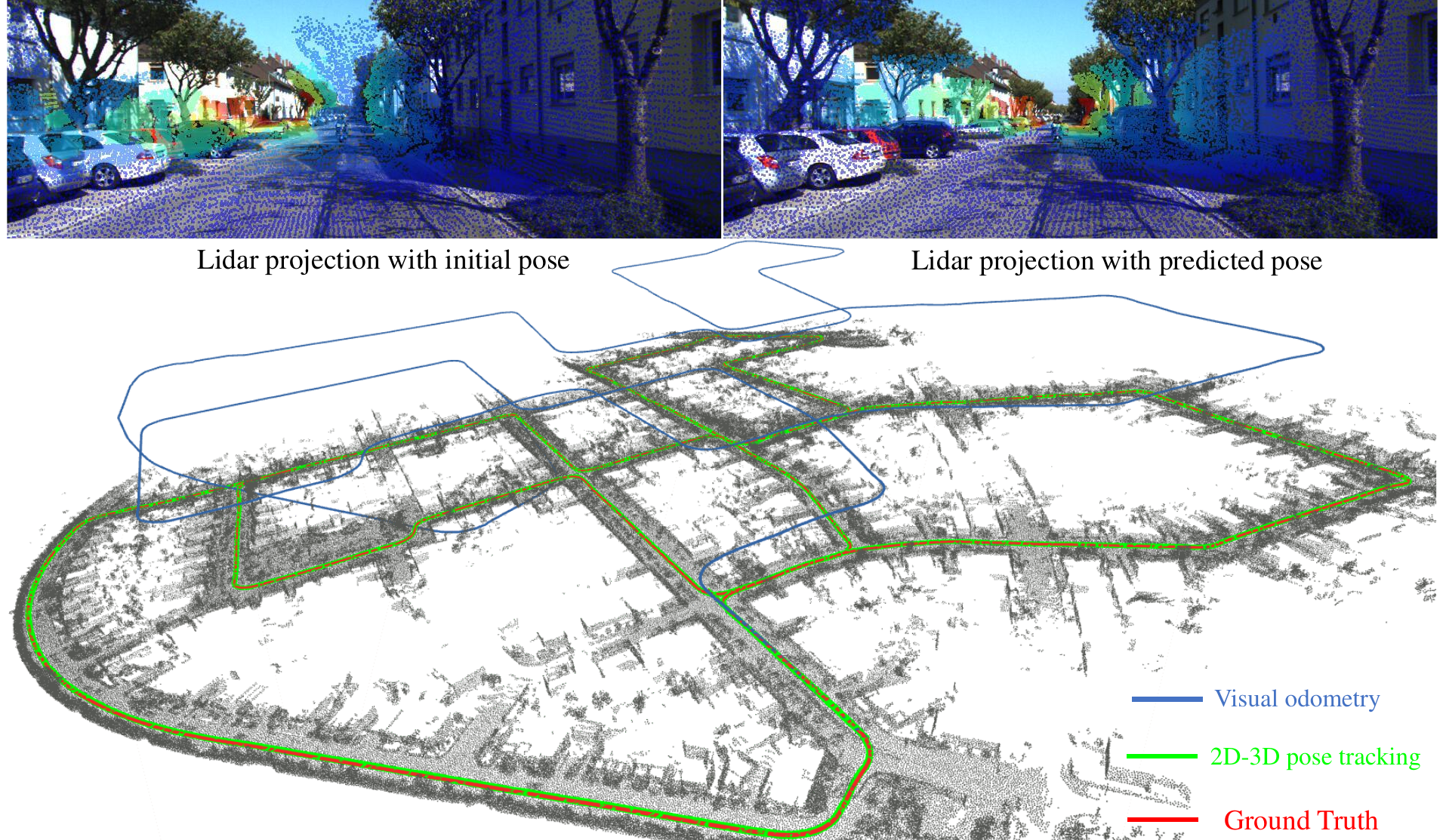}
    \caption{Illustrative examples of the proposed 2D-3D pose tracking and visual odometry on the KITTI 00 sequence. \emph{Top:} The visualization of the LiDAR projection with initial pose and predicted pose. \emph{Bottom:} The top view of the final trajectories of the proposed method and the visual odometry algorithm.} \vspace{-3mm}
    \label{fig:head}
\end{figure}

To alleviate the issue of cross-modal gaps, traditional methods mainly use geometric consistency in 3D or 2D space\cite{caselitz2016monocular, ding2017fusing, zuo2019visual, Liu1990DeterminationOC, Yu2020LineBased2R}. The localization performance highly depends on the accuracy of intermediary ``products" such as reconstructed sparse points or 3D line segments, which lack robustness and generalization capability. With the help of deep learning, the correspondences between LiDAR points and images are established by a cross-modal flow network\cite{Cattaneo2020CMRNetMA, CHEN2022209}, and then the camera poses can be inferred using the PnP solver in a RANSAC loop. Besides, some methods also utilize a neural network to regress the camera poses directly\cite{Cattaneo2019CMRNetCT, Chang2021HyperMapC3, lv2021lccnet}. Despite their remarkable performance, these methods solely estimate the camera pose frame by frame, disregarding the multi-view constraints between adjacent frames, thereby resulting in an unstable pose tracking.

We notice that the bundle adjustment of multi-view image matching can ensure the smoothness of camera pose estimation in classical visual odometry algorithms, and 2D-2D image matching is already a mature technique with traditional handcraft features and learning-based optical flow. Therefore, our intuition is to fuse the LiDAR-image correspondences and image-image flow to establish multi-view visual constraints for further smoothing and improving localization performance.
\IEEEpubidadjcol
In this paper, we propose a novel 2D-3D pose tracking framework that leverages multi-view constraints to achieve accurate pose estimation for consecutive frames. The proposed framework formulates pose estimation as a local optimization problem with an initial pose. Specifically, the initial pose of each frame is from the previous frame during pose tracking. We impose a random disturbance on the ground truth poses to obtain the initial poses during network training. The 3D point clouds are projected onto the 2D plane with rough initial poses to obtain the synthetic depth maps. Then we utilize a hybrid flow estimation network to estimate the image-to-LiDAR depth flows between the adjacent camera frames and a synthetic depth map, as well as the optical flow between the adjacent camera frames. Especially, a cross-modal consistency loss function is devised to incorporate multi-view constraints between the estimated image-to-LiDAR depth flows and the optical flow into the learning process. During pose tracking, we define an objective function consisting of the reprojection error and the cross-modal consistency error, and then optimize the estimated camera poses under the paradigm of a non-linear least square problem. An illustrative example of the proposed framework and visual odometry is shown in Fig. \ref{fig:head}.
% The pose tracking results based on the proposed framework compared with the visual odometry are shown in Fig. \ref{fig:head}.

% In this paper, we propose a novel 2D-3D pose tracking framework tightly coupling the image-to-LiDAR depth flow and the image-to-image flow for 2D-3D pose tracking. In our pipeline, we assume a rough initial camera poses is known, which is obtained by imposing a random disturbance on the ground truth poses during training, and from the previous frame during pose tracking. Based on the initial camera poses, the 3D point clouds are projected to the 2D plane to obtain the synthetic depth maps. Then we train a model to simultaneously estimate the image-to-LiDAR depth flow between the adjacent camera frames and a synthetic depth map, as well as the optical flow between the adjacent camera frames. Especially, a cross-modal consistency loss function is devised to establish multi-view constraint between the estimated image-to-LiDAR depth flow and the optical flow. During pose tracking, we define an objective function consisting of the reprojection error and the cross-modal consistency error, and then optimize the estimated camera poses under the paradigm of BA optimization. The pose tracking results based on the proposed method compared with the visual odometry are shown in Fig. \ref{fig:head}.

\begin{figure*}[htbp]
    \centering
    \includegraphics[width=\linewidth]{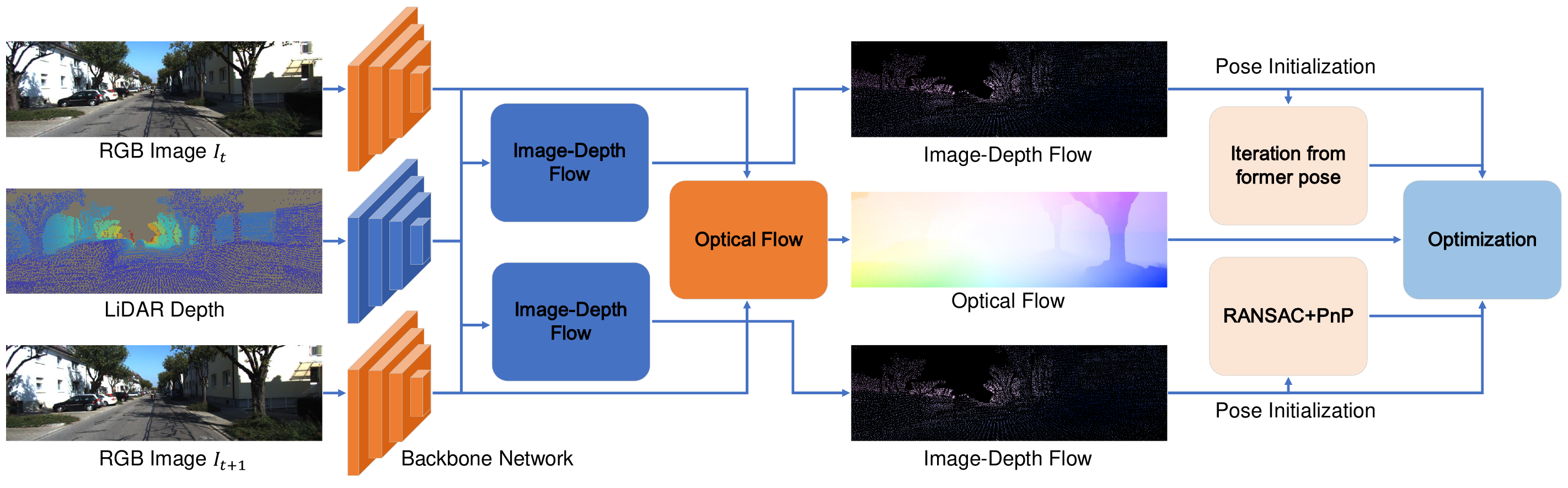}
    \caption{The proposed 2D-3D pose tracking framework. It consists of two main components: the front-end hybrid flow estimation network, and the back-end pose optimization module.} 
    \vspace{-3mm}
    \label{fig:network}
\end{figure*}

Our main contribution can be summarized as:
\begin{itemize}
    \item We propose an effective hybrid flow estimation network that simultaneously estimates image-to-LiDAR depth flow and image-to-image flow between consecutive video frames. We further devise a cross-modal consistency loss function to incorporate multi-view constraints into the learning process.
    % \item An objective function of the back-end nonlinear optimization question is defined, which is solved to optimize the estimated camera poses of the adjacent video frames. 
    \item We introduce a back-end optimization algorithm to smooth and improve localization performance of camera pose tracking in LiDAR maps. The estimated poses corresponding to the consecutive frames are refined together under the paradigm of a non-linear least square problem based on cross-modal consistency.
    % \item Extensive experiments on several LiDAR maps are conducted to show the proposed pose tracking framework is robust and effective.
    \item We conduct extensive experiments on two public datasets (i.e., KITTI and Argoverse) to evaluate the performance. The experimental results demonstrate that the proposed 2D-3D pose tracking framework can achieve more accurate and robust localization than other frame-by-frame pose tracking methods.
\end{itemize}

\section{Related Work}
\label{sec:related_work}

\subsection{Visual-only Pose Tracking}
Cameras provide a wealth of spectral and degenerated geometrical information at a significantly lower cost compared to other competing sensor options. As a result, visual localization plays an important role in modern localization systems for autonomous vehicles. The paradigm of visual localization is to fetch 2D-2D correspondences between two or more 2D images,  typically employing techniques such as optical flow or feature-based matching, and then leverage the epipolar geometry to compute the relative motion. To overcome problems such as scale ambiguity and error accumulation, Triggs \cite{triggs2000bundle} introduces a joint optimization procedure that minimizes the re-projection error between the observed feature points and the estimated points, which is called bundle adjustment optimization. It adjusts the bundle of rays passing through the camera center and the feature points in the 3D world to minimize this error over multiple frames. In addition, some methods \cite{davison2007monoslam, xiao2019real, henawy2020accurate, yu2022tightly} propose to incorporate global information, such as GPS or inertial measurement units (IMUs), to help resolve scale ambiguity and reduce error accumulation during pose tracking. In recent years, with the advent of machine learning and particularly deep learning, learned visual pose tracking methods \cite{li2020deepslam, li2018undeepvo, clark2017vinet, wang2017deepvo} have also emerged. These methods leverage Convolutional Neural Networks (CNNs) to learn features and their associations between frames, thereby estimating camera poses. However, visual-only pose tracking has inevitable pose drift and lacks robustness over challenging conditions such as illumination, weather, and season changes over time.

\subsection{2D-3D Pose Tracking}
Unlike cameras, LiDAR is less susceptible to those visual factors. Therefore, pose tracking based on 2D-3D correspondences between 2D images and offline 3D LiDAR maps holds greater potential in achieving robust localization. Existing approaches focus on solving the cross-modal gaps between 2D images and 3D LiDAR points.
Some researchers propose to extract repeatable points or line features from images and LiDAR points for 2D-3D matching \cite{Feng20192D3DMatchnetLT, wang2021p2, Liu1990DeterminationOC, Yu2020LineBased2R}, which achieve remarkable performance in scenes with rich geometric information.
% Repeatable points or line features are extracted from images and LiDAR points for 2D-3D matching \cite{Feng20192D3DMatchnetLT, wang2021p2, Liu1990DeterminationOC, Yu2020LineBased2R}, which achieve remarkable performance in scenes with rich geometric information.
Other approaches transform data into the same modal at first. 
For example, in certain approaches \cite{caselitz2016monocular, ding2017fusing, zuo2019visual}, sparse 3D points are reconstructed from consecutive camera frames or stereo disparity images captured by a stereo camera. Subsequently, these reconstructed points are matched with the global LiDAR map. The localization performance of these approaches relies on the accuracy of the reconstructed 3D points.
In addition, the 3D reconstruction process is time-consuming and the reconstructed 3D points may not correspond to any points in the LiDAR map.
Other approaches \cite{Cattaneo2020CMRNetMA, wolcott2014visual, stewart2012laps, Cattaneo2019CMRNetCT, Chang2021HyperMapC3, CHEN2022209, lv2021lccnet} first project the LiDAR map onto the 2D plane to obtain synthetic depth maps and then match them with camera images. However, these methods predict the camera pose frame by frame and ignore the constraints between adjacent frames. As a result, they are prone to jitter and encounter error accumulation problems during pose tracking. \cite{nguyen2022calibbd} implicitly utilizes the temporal features between consecutive frames, but still predicts the camera poses frame by frame. In this paper, we propose a novel 2D-3D pose tracking framework which integrates the multi-view constraints between consecutive frames during the network training and inference. Unlike the aforementioned methods, we regard the poses of multiple frames as a whole and optimize them together under the proposed cross-modal consistency, thus can well handle the cross-modal differences and multi-view constraints.

\section{Proposed Method}
\label{sec:method}
% Assuming a pose tracking paradigm based on 2D-3D correspondences: At first, the camera pose of the first frame is initialized using GPS or other global positioning methods. Subsequently, for each captured camera frame, the camera pose corresponding to the previous frame serves as the initial pose of the current one. Based on this pose, a fixed-size point cloud is segmented from the global LiDAR map. Finally, the 2D-3D correspondences between the current camera frame and the point cloud are estimated, after which the corresponding camera pose is calculated using the PnP solver. Previous work\cite{CHEN2022209} has verified this paradigm's feasibility for pose tracking. However, they only estimate the camera poses frame by frame, which ignores the multi-view constraint between the adjacent frames, making the pose tracking unstable and prone to failure when encountering some scenarios. 

The general frame-by-frame 2D-3D pose tracking framework adheres to the paradigm of local optimization based on an initial pose, which can be formulated as follows: In the beginning, the camera pose of the first frame is initialized using GPS. For each subsequent camera frame, the pose of the previous frame serves as the initial estimation. Based on this initial pose, a fixed-size point cloud is segmented from the global LiDAR map. Then, the 2D-3D correspondences between the camera frame and the segmented point cloud are estimated. Finally, the corresponding camera pose is calculated using the PnP solver. Previous work\cite{Cattaneo2019CMRNetCT, Chang2021HyperMapC3, CHEN2022209} has confirmed the feasibility of this paradigm for pose tracking. However, it has also highlighted the limited robustness in some scenarios, such as environments with extreme degeneracy. 
The unstable camera localization motivates us to incorporate multi-view constraints between adjacent camera frames into the 2D-3D pose tracking framework.

Our proposed 2D-3D pose tracking framework is shown in Fig. \ref{fig:network}. 
% Similar to the frame-by-frame pose tracking framework, the pose of the first frame is initialized using GPS or other global positioning methods. However, our framework differs in that we simultaneously estimate the 6-DoF poses for two adjacent camera frames based on a hybrid flow estimation network.
Our key insight is to simultaneously estimate the 6-DoF poses for two adjacent frames based on a hybrid flow estimation network. For each group of two adjacent frames, we initialize the poses as the same and from the estimation of previous continuous frames. Then the global LiDAR maps are cropped with a fixed size centered at the initial poses. By using the hybrid 2D-3D and 2D-2D flow network, we can obtain stable 2D-3D correspondences. PnP solver is then utilized to get the estimated poses.
Finally, these poses are further refined using a back-end optimization algorithm that incorporates multi-view constraints. Subsequently, the initial pose of the current time step is updated with the estimated pose of the current frame, while the estimated pose of the next frame serves as the initial pose for the next time step.
% In this work, we propose another pose tracking paradigm based on 2D-3D correspondences: Similarly, the initial pose of the first frame is obtained using GPS or other localization methods such as image retrieval. The difference is that we estimate the 2D-3D correspondences between adjacent two camera images and the segmented point cloud simultaneously, as well as the optical flow between the two images. After solving the corresponding camera poses, a back-end optimization algorithm is utilized to optimize the estimated poses based on the multi-view constraint. Finally, the initial pose of the current time is updated with the estimated current frame's pose, and the estimated next frame's pose is used as the initial pose of the next time.

% In the following, we first provide a general description of our global matching formulation, and then present a Transformer-based framework to realize it.
In the following, we first introduce the front-end flow estimation network, and then describe the details of the proposed cross-modal consistency loss function. Finally, we present the back-end optimization algorithm.

\subsection{Hybrid Flow Estimation Network}
% Our proposed 2D-3D pose tracking framework is shown in Fig. \ref{fig:network}. 
% Considering the efficiency and computation burden, we use the projection-based method to match the RGB images and the point clouds. 
\subsubsection{Depth Map Generation}
To obtain the image-to-LiDAR depth flow using the flow estimation network, we first need to project the 3D LiDAR point clouds to the 2D plane to generate the synthetic depth maps. The projection process follows the pinhole camera model which can be described as:
\begin{equation}
\left(\begin{array}{l}
u \\
v \\
1
\end{array}\right)=\frac{1}{Z}\left(\begin{array}{ccc}
f_x & 0 & c_x \\
0 & f_y & c_y \\
0 & 0 & 1
\end{array}\right)\left(\begin{array}{l}
X \\
Y \\
Z
\end{array}\right) \triangleq \frac{1}{Z} \boldsymbol{K} P
\end{equation}
% where $(X \quad Y \quad Z)^{T}$ and $P$ both represent the coordinates of the 3D points in the camera coordinate system. $(u \quad v \quad 1)^{T}$ is the homogeneous coordinates of the projection in the pixel coordinate system. $f_x, f_y, c_x, c_y$ are the intrinsic camera parameters and they construct the camera matrix $\boldsymbol{K}$. In addition, the occluded 3D points on the generated depth map are removed based on the occlusion removal scheme proposed by \cite{Pintus2011RealtimeRO}.
where $(X \quad Y \quad Z)^{T}$ and $P$ represent the coordinates of the 3D points in the camera coordinate system. $(u \quad v \quad 1)^{T}$ represents the homogeneous coordinates of the projection in the pixel coordinate system. The intrinsic camera parameters, $f_x, f_y, c_x, c_y$, form the camera matrix $\boldsymbol{K}$. Additionally, the occlusion removal scheme proposed by \cite{Pintus2011RealtimeRO} is applied to eliminate occluded 3D points on the generated depth map.

\subsubsection{Flow Estimation}
After obtaining the synthetic depth map, we utilize a hybrid flow estimation network to predict the image-to-LiDAR depth flow and the optical flow between the consecutive camera frames simultaneously. The hybrid network mainly consists of three parts: the image-to-LiDAR depth flow estimation networks $F_{c}$ and $F_{n}$, and the optical flow estimation network $F_{i}$. We use I2D-Loc\cite{CHEN2022209} as the backbone to predict the image-to-LiDAR depth flow and RAFT\cite{Teed2020RAFTRA} as the backbone to predict the optical flow because they both excel at their respective tasks. 
% Next, we use the proposed flow estimation network to predict the image-to-depth flow between the current RGB image and the synthetic depth map, another image-to-depth flow between the next RGB image and the synthetic depth map, and the optical flow between the two adjacent RGB images respectively. Correspondingly, the flow estimation network consists of three parts: the image-to-depth flow estimation networks $F_{c}$ and $F_{n}$, and the optical flow estimation network $F_{i}$. We use I2D-Loc\cite{CHEN2022209} as the backbone to predict the image-to-depth flow and RAFT\cite{Teed2020RAFTRA} as the backbone to predict the optical flow because they have remarkable performance on these tasks. 

The ground truth image-to-LiDAR depth flow is generated by computing the distance between the LiDAR depth maps projected based on the initial pose and the ground truth pose. Given the initial pose $\boldsymbol{T}_{\text{init}}$, the ground truth image-to-LiDAR depth flow for the current frame can be computed using the following formulation:
\begin{equation}
    {[\Delta u, \Delta v]_{\text {cur2depth}}=h\left(P_w, \hat{\boldsymbol{T}}_{\text {cur}}\right)-h\left(P_w, \boldsymbol{T}_{\text {init}}\right)}
\end{equation}
\begin{equation}
    h(P, \boldsymbol{T}) \triangleq \boldsymbol{K T} P \label{formula:projection}
\end{equation}
where $[\Delta u, \Delta v]_{\text {cur2depth}}$ is the calculated ground truth image-to-LiDAR depth flow between the current camera frame and the synthetic depth map. $\hat{\boldsymbol{T}}_{\text{cur}}$ and $\boldsymbol{T}_{\text{init}}$ are the ground truth pose and the initial pose, respectively. $P_w$ is the coordinate of the corresponding 3D point cloud in the world coordinate system. $h$ represents the camera projection function that first transforms 3D points from the world coordinate system to the camera coordinate system and then projects them to the 2D plane. Besides, the ground truth image-to-LiDAR depth flow for the next frame can be calculated based on a similar formulation:
\begin{equation}
    {[\Delta u, \Delta v]_{\text {next2depth}}=h\left(P_w, \hat{\boldsymbol{T}}_{\text {next}}\right)-h\left(P_w, \boldsymbol{T}_{\text {init}}\right)}
\end{equation}
where $[\Delta u, \Delta v]_{\text {next2depth}}$ represents the calculated ground truth image-to-LiDAR depth flow between the next camera frame and the synthetic depth map. $\hat{\boldsymbol{T}}_{\text{next}}$ is the ground truth pose of the next camera frame. 

% However, to reduce the computation burden, we actually segment fixed-size point cloud from the global LiDAR map according to the ground truth pose of the current frame during training.
% \begin{figure*}[htbp]
%     \centering
%     \includegraphics[width=\linewidth]{figures/Consistency.pdf}
%     \caption{Visualization of the flow estimation.} 
%     \vspace{-3mm}
%     \label{fig:consist}
% \end{figure*}

\subsection{Loss Function}
% Regarding network supervision, we consider the flow error and then account for the cross-modal consistency between the predicted image-to-LiDAR depth flow and the predicted optical flow.
Regarding the network supervision, we first consider the masked average endpoint error (EPE) between the predicted image-to-LiDAR depth flow and the ground truth, defined as follows:
% The flow error is measured as the average endpoint error between the predicted image-to-LiDAR depth flow and the ground truth, defined as follows:
\begin{equation}
    L_{\text{epe}}=\frac{\sum g(u, v)\left\|f_{\text{pre}}(u, v)-f_{\text{gt}}(u, v)\right\|_2}{\sum g(u,v)} 
\end{equation}
\begin{equation}
    g(u, v)=\left\{\begin{array}{l}1, f_{\text{gt}} \neq 0 \\
0, \text {otherwise}
\end{array}\right.
\end{equation}
% where $f_{\text{pre}}$ represents the predicted image-to-LiDAR depth flow. $f_{\text{gt}}$ is the ground truth image-to-LiDAR depth flow. 
where $f_{\text{pre}}$ and $f_{\text{gt}}$ represent the predicted and ground truth image-to-LiDAR depth flow, respectively.
$g(u,v)$ identifies the valid pixels in the ground truth image-to-LiDAR depth flow. 
Due to the superior accuracy achieved by the same modal image optical flow network compared to the image-to-LiDAR depth flow network, we load a pre-trained optical flow model and keep its parameters fixed during network training.
% Since the optical flow estimation network\cite{Teed2020RAFTRA} achieves higher accuracy compared to the image-to-LiDAR depth flow estimation network\cite{CHEN2022209}, we load the pre-trained model and keep the parameters of the optical flow estimation network fixed during network training.

\begin{figure}[htbp]
    \centering
    \includegraphics[width=\linewidth]{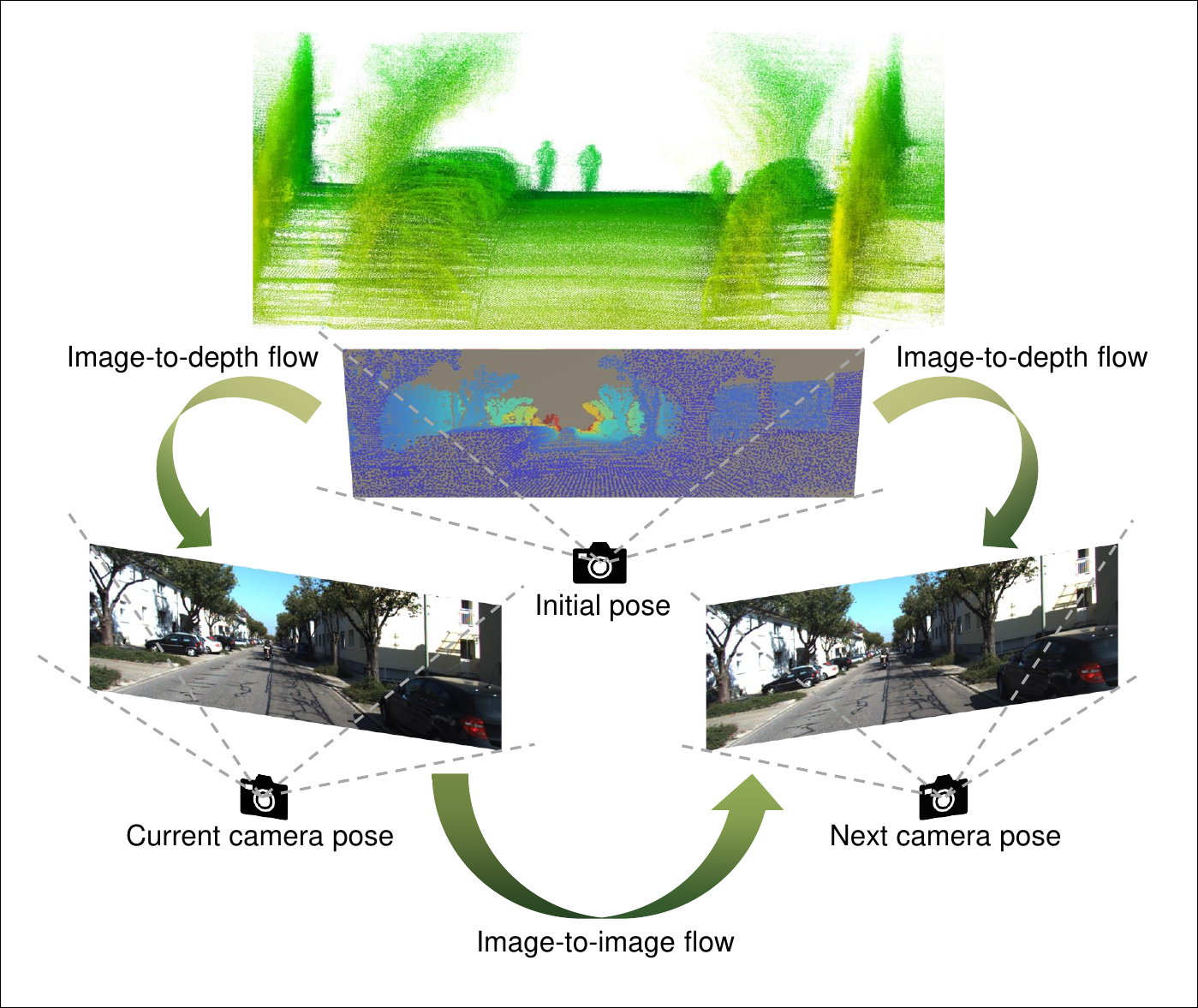}
    \caption{Diagram of the cross-modal consistency.} 
    \vspace{-3mm}
    \label{fig:consist_display}
\end{figure}

\begin{figure}[htbp]
    \centering
    \includegraphics[width=\linewidth]{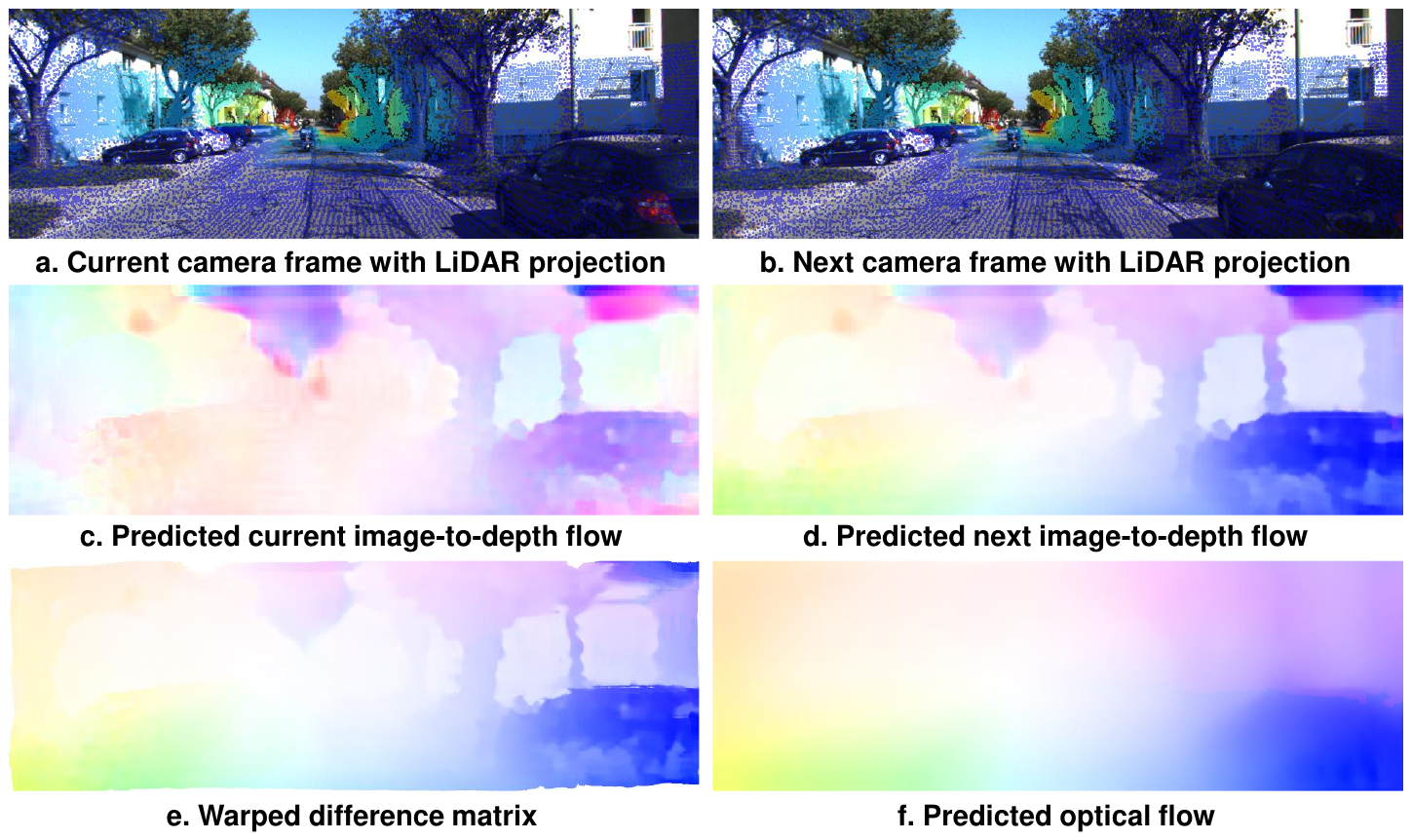}
    \caption{Visualization of the flow estimation. \emph{a:} LiDAR projection overlay on the current camera image. \emph{b:} LiDAR projection overlay on the next camera image. \emph{c:} Predicted image-to-LiDAR depth flow between the current image and the LiDAR projection. \emph{d:} Projected image-to-LiDAR depth flow between the next image and the LiDAR projection. \emph{e:} Warped difference matrix between the predicted image-to-depth flows. \emph{f:} Predicted optical flow between the two camera images.} 
    \vspace{-3mm}
    \label{fig:consist}
\end{figure}

% The second part of the loss function is the proposed cross-modal consistency loss, which is devised as follows:
% Additionally, to couple the predicted image-to-LiDAR depth flows and the predicted optical flow during the learning process, we propose a cross-modal consistency-based loss function, which is formulated as follows:
Additionally, to incorporate the multi-view constraints between two adjacent camera frames during the network training, we propose a cross-modal consistency-based loss function, which is formulated as follows:
\begin{equation}
L_{\text {consist}}=\left\|w\left(f_{\text {pre}}^{\text{n2d}}-f_{\text{pre}}^{\text{c2d}}, f_{\text{pre}}^{\text{c2d}}\right)-f_{\text {pre}}^{\text{c2n}}\right\|_2
\end{equation}
where $f_{\text{pre}}^{\text{n2d}}$ represents the predicted image-to-LiDAR depth flow between the next camera frame and the synthetic depth map. Similarly, $f_{\text{pre}}^{\text{c2d}}$ represents the predicted image-to-LiDAR depth flow for the current camera frame. $f_{\text {pre}}^{\text{c2n}}$ is the predicted optical flow between the adjacent camera frames. $w$ represents the warping operation. 

The cross-modal consistency, as shown in Fig. \ref{fig:consist_display}, explicitly describes the relationship between the adjacent camera frames and the synthetic depth map.
Specifically, the image-to-depth flow represents the 2D-3D correspondences between the camera frame and depth map. Fig. \ref{fig:consist}c and \ref{fig:consist}d visualize the predicted flows corresponding to the current and next frame, respectively. The image-to-image flow (i.e. optical flow) represents the 2D-2D correspondences between two adjacent camera frames, which is shown in Fig. \ref{fig:consist}f. By calculating the difference between the predicted image-to-depth flows, we will obtain the equivalent image-to-image flow between the camera frames. As a result, the error between the equivalent and predicted image-to-image flow constructs the multi-view constraints between these predictions.
% In addition, since the predicted image-to-LiDAR depth flow represents the displacement field from the synthetic depth map to the image, the difference matrix needs to be warped based on the current image-to-LiDAR depth flow to be equivalent to the predicted optical flow.
% In addition, since the predicted image-to-LiDAR depth flow represents the displacement field from the synthetic depth map to the image, the direct difference between two image-to-LiDAR depth flows isn't equivalent to the 2D-2D correspondences between the corresponding frames.
In addition, it's important to note that the predicted image-to-LiDAR depth flow represents the displacement field from the synthetic depth map to the image. Hence, the difference between the image-to-LiDAR depth flows of two adjacent frames is not equivalent to the image-to-image flow between them. We need to warp the calculated difference matrix based on the image-to-LiDAR depth flow of the first frame to obtain the final equivalent optical flow.

The final loss function is the sum of the above loss functions:
\begin{equation}
    L = L_{\text{epe}}^{\text{cur}} + L_{\text{epe}}^{\text{next}} + L_{\text{consist}}
\end{equation}
where $L_{\text{epe}}^{\text{cur}}$ is the average endpoint error of the predicted current image-to-depth flow. $L_{\text{epe}}^{\text{next}}$ is the average endpoint error of the predicted next image-to-depth flow.

% Above all, we utilize two kinds of loss functions to supervise the network training: the average endpoint error and the cross-modal consistency error. The former is the constraint used for traditional optical flow estimation tasks. The latter establishes the multi-view constraint between the adjacent camera frames.
In summary, we employ two loss functions to guide the network training: the average endpoint error and the cross-modal consistency loss function. The average endpoint error serves as a conventional constraint commonly used in optical flow estimation tasks. The cross-modal consistency error introduces the multi-view constraints between adjacent camera frames, capturing the relationship between different views.

\subsection{Multi-View Constraints-based Back-End Optimization}
During pose tracking, as previously mentioned, we utilize the trained hybrid flow estimation network to simultaneously predict the image-to-LiDAR depth flows between the adjacent camera frames and the synthetic depth map, as well as the optical flow between the camera frames. Then the camera poses corresponding to the two camera frames are solved using the PnP solver according to the prediction. 
Besides, we define an energy function to further optimize the solved poses, as:
% To make the localization smoothing and stable, we design an objective function to optimize the calculated camera poses based on the multi-view constraint as follows:
\begin{equation}
    \boldsymbol{T}_{\text{cur}}^*, \boldsymbol{T}_{\text{next}}^*=\argmin _{\boldsymbol{T}_{\text{cur}}, \boldsymbol{T}_{\text{next}}} ( E_{\text{consist}} + E_{\text{reproj}}^{\text{cur}} + E_{\text{reproj}}^{\text{next}}),
\end{equation}
where $\boldsymbol{T}_{\text{cur}}$ and $\boldsymbol{T}_{\text{next}}$ represent the predicted camera poses corresponding to the current and next frames, respectively. $E_{\text{consist}}$ is the cross-modal consistency error and formulated as follows:
\begin{equation}
    E_{\text{consist}} = \left\Vert h(P_w,\boldsymbol{T}_{\text{next}})-h(P_w,\boldsymbol{T}_{\text{cur}})-f^{\text{c2n}}_{\text{pre}}\right\Vert_{2}
\end{equation}
where $h$ is the camera projection function and defined as Eq. (\ref{formula:projection}). $f^{\text{c2n}}_{\text{pre}}$ is the predicted optical flow.
% According to this formula, we first project the 3D points to the 2D plane based on the predicted poses, and then calculate the difference matrix between the projections to obtain the equivalent optical flow.
Based on this formula, we begin by projecting the 3D points onto the 2D plane using the predicted poses. Subsequently, we compute the difference matrix between the projections to derive the equivalent optical flow. Therefore, $E_{\text{consist}}$ represents the distance between the equivalent and predicted optical flows. 
Additionally, $E_{\text{reproj}}^{\text{cur}}$ and $E_{\text{reproj}}^{\text{next}}$ both represent the reprojection error of the solved camera poses:
\begin{equation}
     E_{\text{reproj}}^{\text{cur}} = \left\Vert h(P,\boldsymbol{T}_{\text{cur}})-X_{\text{cur}}\right\Vert_{2}
\end{equation}
\begin{equation}
    E_{\text{reproj}}^{\text{next}} = \left\Vert h(P,\boldsymbol{T}_{\text{next}})-X_{\text{next}}\right\Vert _{2}
\end{equation}
where $X_{\text{cur}}$ and $X_{\text{next}}$ represent the 2D coordinates of pixels with valid depth values in the depth maps that have been warped using the predicted image-to-LiDAR depth flows.
% the 2D coordinates of the pixels with valid depth values in the depth maps warped based on the predicted image-to-depth flows.

Based on the formulated energy function, we employ the least square method to determine the optimal camera poses $\boldsymbol{T}_{\text{cur}}^*$ and $\boldsymbol{T}_{\text{next}}^*$ that minimize the function value. The obtained $\boldsymbol{T}_{\text{cur}}^*$ replaces the initial pose $\boldsymbol{T}_{\text{init}}$ as the final pose for the current frame, while $\boldsymbol{T}_{\text{next}}^*$ is utilized to initialize the subsequent localization process.
\section{Experiments}
% In this section, we conduct extensive experiments on the KITTI odometry dataset\cite{Geiger2013VisionMR} to verify the effectiveness of the proposed method.
In this section, we conduct extensive experiments on two public datasets to evaluate the performance of our proposed method.

\begin{table*}
    \caption{Performance comparison of the flow estimation network under different setups on the KITTI dataset. ``C", ``N", ``T", and ``L" represent the current image-to-LiDAR depth flow estimation network $F_c$, the next image-to-LiDAR depth flow estimation network $F_n$, additional training, and the cross-modal consistency loss function, respectively. In (e) and (f), the first value represents the performance of the network $F_c$, while the second value represents the performance of the network $F_n$.}
    \centering
    \begin{tabular}{c|cccc|cc|cc|c}
    \hline
    \multirow{2}{*}{Case} & \multicolumn{4}{c|}{Module}  & \multicolumn{2}{c|}{Mean Error} &\multicolumn{2}{c|}{Median Error} &\multirow{2}{*}{Fail[\%] $\downarrow$}\\ 
    & C & N & T & L & Rot.[$^\circ$] $\downarrow$ & Transl.[cm] $\downarrow$ & Rot.[$^\circ$] $\downarrow$ & Transl.[cm] $\downarrow$ \\
    \hline
    Initial pose & & & & & $\approx9.6726$ & $\approx182.8381$  & $\approx9.9033$  & $\approx187.3079$  & -\\
    (a) & \checkmark & & &  & 0.8619 & 24.6955 & 0.6900 & 17.6687 & 1.81 \\
    (b) & & \checkmark & &  & 2.5284 & 75.4198 & 1.8364 & 63.5510 & 16.65 \\
    (c) & \checkmark & & \checkmark &  & \bf{0.8375} & \bf{22.4241} & \bf{0.6777} & \bf{15.3426} & \bf{1.61} \\
    (d) & & \checkmark & \checkmark &  & \bf{2.4225} & 74.0713 & 1.8184 & 62.6881 & 16.67 \\
    (e) & \checkmark & \checkmark & \checkmark &  & 0.8640/2.4801 & 23.7909/75.5822 & 0.6978/1.8511 & 16.0583/63.5162 & 1.61/16.93 \\
    (f) & \checkmark & \checkmark & \checkmark & \checkmark & 0.9051/2.4842 & 26.5121/\bf{71.8964} & 0.7168/\bf{1.7970} & 17.8525/\bf{61.0808} & 1.78/\bf{15.90} \\
    \hline
    \end{tabular}
    \label{tab:1}
\end{table*}

\subsection{Dataset and Evaluation Metrics}
\subsubsection{Dataset}
We conduct experiments on KITTI\cite{Geiger2013VisionMR} and Argoverse\cite{Chang2019Argoverse3T} datasets.
KITTI is commonly used in the field of autonomous driving. It consists of 22 sequences. We use sequences 03, 05, 06, 07, 08, and 09 as the training set and sequences 00 and 10 as the validation set. Argoverse offers more complex driving scenarios, albeit with shorter odometry sequences. We use the sequences train1, train2, and train3 as the training set, and sample several sequences with less noise as the validation set.

% To obtain the global LiDAR map which is not provided by the dataset, we first aggregate all scans according to the ground truth poses and then down-sample the aggregation at a resolution of 0.1m to save storage space.
To acquire the global LiDAR map, unavailable in the datasets, we begin by aggregating all scans based on the ground truth poses. Subsequently, to conserve storage space, we down-sample the aggregated map at a resolution of 0.1m. 
During training, we segment the point cloud for each frame by centering it around the ground truth pose and extending it 100m forward, 10m backward, and 25m to the left and right. During pose tracking, the coverage of the segmented point cloud remains fixed, but it is centered on the initial pose of the current frame being tracked.
% During training, for each frame, we segment the corresponding point cloud centering at the ground truth pose, extending 100m forward, 10m backward, and 25m left and right, respectively. During pose tracking, the coverage of the segmented point cloud keeps fixed, however, the point cloud is centered on the initial pose of the corresponding current frame instead. 

\subsubsection{Evaluation Metrics}
We evaluate the performance of the proposed method in two ways. The first one is to evaluate the localization accuracy frame by frame using the mean and median errors of the predicted poses. Following \cite{Cattaneo2020CMRNetMA}, \cite{CHEN2022209}, we assume the position errors larger than four meters as a \emph{failure}.
The second one is to evaluate the pose tracking performance on the global LiDAR maps using the average trajectory error (ATE) and the relative pose error (RPE), which are defined in \cite{zhang2018tutorial} and commonly used for quantitative trajectory evaluation.

\subsection{Experimental Setup}
% Considering the sizes of the RGB images in different sequences differ, we crop all the RGB images to $960\times320$ for network training.
Accounting for the varying sizes of RGB images across different sequences, we uniformly crop all RGB images to a resolution of $960\times320$ during network training.
% As mentioned above, the proposed network consists of three parts: the current image-to-LiDAR depth flow estimation network $F_c$, the next image-to-LiDAR depth flow estimation network $F_n$, and the optical flow estimation network $F_i$.
As mentioned above, the proposed hybrid network comprises three components: the current image-to-LiDAR depth flow estimation network $F_c$, the next image-to-LiDAR depth flow estimation network $F_n$, and the optical flow estimation network $F_i$.
% We use I2D-Loc\cite{CHEN2022209} as the backbone of $F_c$ and $F_n$, and use RAFT-S\cite{Teed2020RAFTRA} as the backbone of $F_i$ due to the limited graphics memory.
We employ I2D-Loc\cite{CHEN2022209} as the backbone for $F_c$ and $F_n$, while for $F_i$, we utilize RAFT-S\cite{Teed2020RAFTRA} as the backbone due to memory limitations.
% Especially, we load the pre-trained model provided by \cite{CHEN2022209, Teed2020RAFTRA} for convenience.
In particular, we initialize $F_c$ and $F_i$ by loading the pre-trained models provided by \cite{CHEN2022209, Teed2020RAFTRA}.
As for $F_n$, we first train it singly for 100 epochs based on the same experimental setup in \cite{CHEN2022209}.
% After that, we load all the pre-trained model weights and train $F_c$, $F_n$, and $F_i$ together for 50 more epochs.
Subsequently, we load all the pre-trained model weights and conduct joint training for additional 50 epochs on $F_c$, $F_n$, and $F_i$ using the proposed cross-modal consistency loss function.
% The learning rate, weight decay, and training batch size are set to $4\times10^{-6}$, $1\times10^{-4}$, and 2, respectively. We use \emph{MultiStepLR} as the learning rate scheduler and the learning rate decays to one-tenth of the previous one at the 10th and 30th epoch.
The learning rate, weight decay, and training batch size are set to $4\times10^{-6}$, $1\times10^{-4}$, and 2, respectively. We employ the \emph{MultiStepLR} learning rate scheduler, which reduces the learning rate to one-tenth of the previous value at the 10th and 30th epochs.
% All the experiments are conducted on two NVIDIA RTX 3090 Ti GPUs.
The joint training is performed using two NVIDIA RTX 3090 Ti GPUs, while the other experiments are conducted on a single NVIDIA RTX 3090 Ti GPU.

\begin{table*}
    \caption{Pose tracking performance comparison on KITTI sequence 00. ``Complete" indicates that the pose tracking process ends without any interruptions caused by localization failure. The localization errors are calculated until being interrupted.}
    \centering
    \begin{tabular}{c|c|cc|cc|c|c|c}
    \hline
    \multirow{2}{*}{Case} &
    \multirow{2}{*}{Method} &
    \multicolumn{2}{c|}{Translation Error} & 
    \multicolumn{2}{c|}{Rotation Error} & 
    \multirow{2}{*}{Complete} &
    \multirow{2}{*}{Param[M]} &
    \multirow{2}{*}{Time[ms]}\\ 
    & & Mean.[cm] $\downarrow$ & Std.[cm] $\downarrow$ & Mean.[$^\circ$] $\downarrow$ & Std.[$^\circ$] $\downarrow$ & &\\
    \hline
    (a) & CMRNet\cite{Cattaneo2019CMRNetCT} & 205 & 107 & 126 & 39 & \checkmark & 37.1 & 25\\
    (b) & I2D-Loc\cite{CHEN2022209} & 21 & 10 & 0.46 & 0.30 &  & 6.3 & 176\\
    (c) & VO & 162 & 63 & 4.95 & 2.34 & \checkmark & - & 219\\
    (d) & I2D-VO & 15 & 21 & 0.48 & 0.64 & \checkmark & 6.3 & 180\\
    (e) & Ours w/o Optim & 14 & 17 & 0.50 & 0.53 &  & 13.6 & 374\\
    (f) & Ours w/  Optim & 14 & 17 & 0.49 & 0.54 & \checkmark & 13.6 & 588\\
    % (f) & Ours-S w/  Optim & 18 & 31 & 0.51 & 0.69 & & 5.7 & 521 \\ 
    \hline
    \end{tabular}
    \label{tab:2}
\end{table*} 

\begin{figure*}[htbp]
    \centering
    \includegraphics[width=\linewidth]{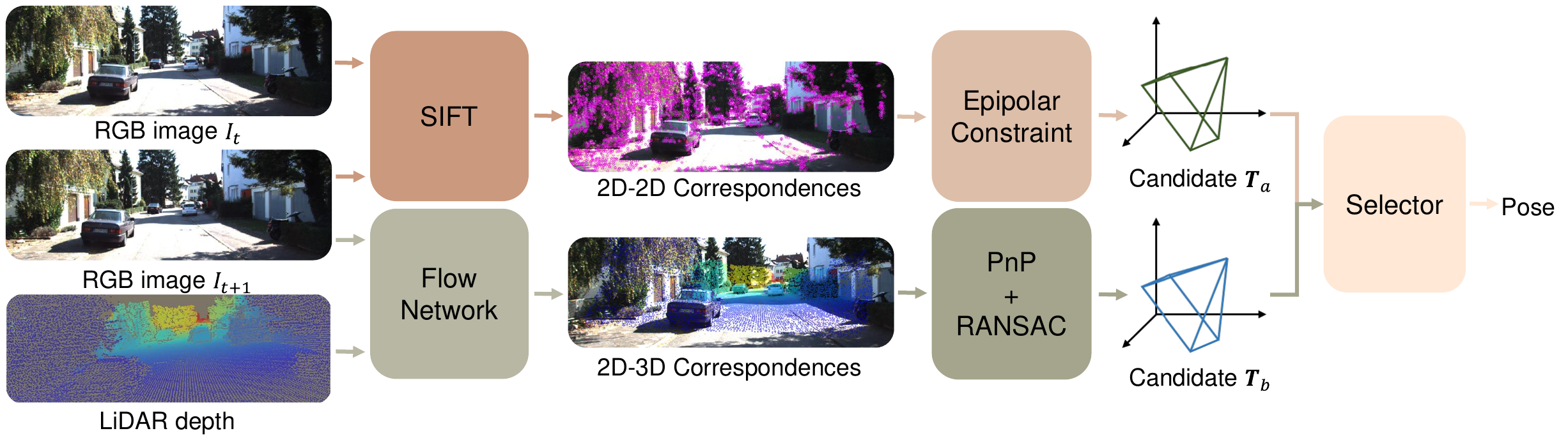}
    \caption{2D-3D pose tracking framework that loosely couples 2D-3D correspondences and 2D-2D correspondences.} 
    \vspace{-3mm}
    \label{fig:loose}
\end{figure*}

% 	\caption{Performance comparison of different methods for pose tracking 
%  % Performance of different methods for tracking localization in LiDAR point cloud maps corresponding to the 00 sequence of the KITTI dataset. It should be noted that the mean and standard deviation of the translation error and rotation error of the evaluation index here calculate the mean and standard deviation of the pose error at each moment. "Whether the route is completed" refers to whether the method can be used to locate in the point cloud map from the beginning to the end without experiencing the situation of tracking interruption caused by positioning failure.

\subsection{Ablation Study}
% In this section, we evaluate the effectiveness of our proposed method under varying setups on the KITTI 00 sequence. The results are shown in Table \ref{tab:1}, where we report the mean and median errors of the predicted poses as evaluation metrics.
In this section, we evaluate the effectiveness of the multi-view constraints during the training process. The experimental results are shown in Table \ref{tab:1}, where we report the mean and median errors of the calculated poses based on the predicted flows as evaluation metrics.

% The performance of the single image-to-LiDAR depth flow estimation network, loaded with pre-trained weights, are displayed in Table \ref{tab:1}(a) and (b).
The performance of the single image-to-LiDAR depth flow estimation network for current and next frames using the official I2D-Loc model is displayed in Table \ref{tab:1}(a) and (b).
The error range of the initial poses for the network $F_n$ is greater than that for the network $F_c$, resulting in a relatively higher localization accuracy for the network $F_c$. 
After conducting an additional 50 epochs of training for each network individually in Table \ref{tab:1}(c) and (d), the pose errors of the two networks both decrease by a little margin. 
Then when the networks $F_c$ and $F_n$ are trained jointly for another 50 epochs, Table \ref{tab:1}(e) shows that the localization accuracy for each network hasn't been improved. The above experiments imply an inherent conflict between the network $F_c$ and $F_n$ due to their varying error ranges for initial poses.
In Table \ref{tab:1}(f), we utilize the proposed cross-modal consistency loss function to incorporate the multi-view constraints during the training process. The experimental result shows that the localization performance of the network $F_c$ degrades a little, but the pose error of the network $F_n$ decreases by a large margin. This outcome demonstrates that the proposed cross-modal consistency-based loss bridges the gaps between the predicted 2D-3D correspondences of adjacent frames.

\subsection{Results Analysis}

Table \ref{tab:2} gives the quantitative results of our method evaluated on KITTI sequence 00.
In addition to the frame-by-frame 2D-3D pose tracking methods CMRNet\cite{Cattaneo2019CMRNetCT} and I2D-Loc\cite{CHEN2022209}, we also compare our method with the traditional visual odometry algorithm and a devised 2D-3D pose tracking framework that loosely couples the visual odometry algorithm with I2D-Loc, referred to as I2D-VO.
As illustrated in Fig. \ref{fig:loose}, this framework utilizes the visual odometry algorithm and I2D-Loc to generate pose candidates based on 2D-2D and 2D-3D correspondences, respectively. Subsequently, the final pose is selected from these candidates based on predefined thresholds.
% As shown in Fig. \ref{fig:loose}, we use feature matching algorithm (such as SIFT\cite{Lowe1999ObjectRF}) to obtain the 2D-2D correspondences between the current RGB image $I_{t+1}$ and the reference RGB image $I_{t}$, and then solve the corresponding camera pose based on the epipolar constraint. Simultaneously, we use I2D-Loc\cite{CHEN2022209} to predict the 2D-3D correspondences between the current RGB image $I_{t}$ and the LiDAR depth map generated based on an initial pose, and then solve the corresponding camera pose using the PnP solver. Subsequently, according to a pre-defined threshold, the final pose is identified from the pose candidates. 

% The results are shown in Table \ref{tab:2}. Unlike the proposed pose tracking network based on the multi-view constraint, we also devise a pose tracking method loosely coupling 2D-3D correspondences and 2D-2D correspondences, the framework of which is shown in Fig. \ref{fig:soft}. We use feature matching algorithm (such as SIFT\cite{Lowe1999ObjectRF}) to obtain the 2D-2D correspondences between the current RGB image $I_{t+1}$ and the reference RGB image $I_{t}$, and then solve the corresponding camera pose based on the epipolar constraint. Simultaneously, we use I2D-Loc\cite{CHEN2022209} to predict the 2D-3D correspondences between the current RGB image $I_{t}$ and the LiDAR depth map generated based on an initial pose, and then solve the corresponding camera pose using the PnP solver. Subsequently, according to a pre-defined threshold, the final pose is identified from the pose candidates. 

\begin{figure}[htbp]
    \centering
    \includegraphics[width=\linewidth]{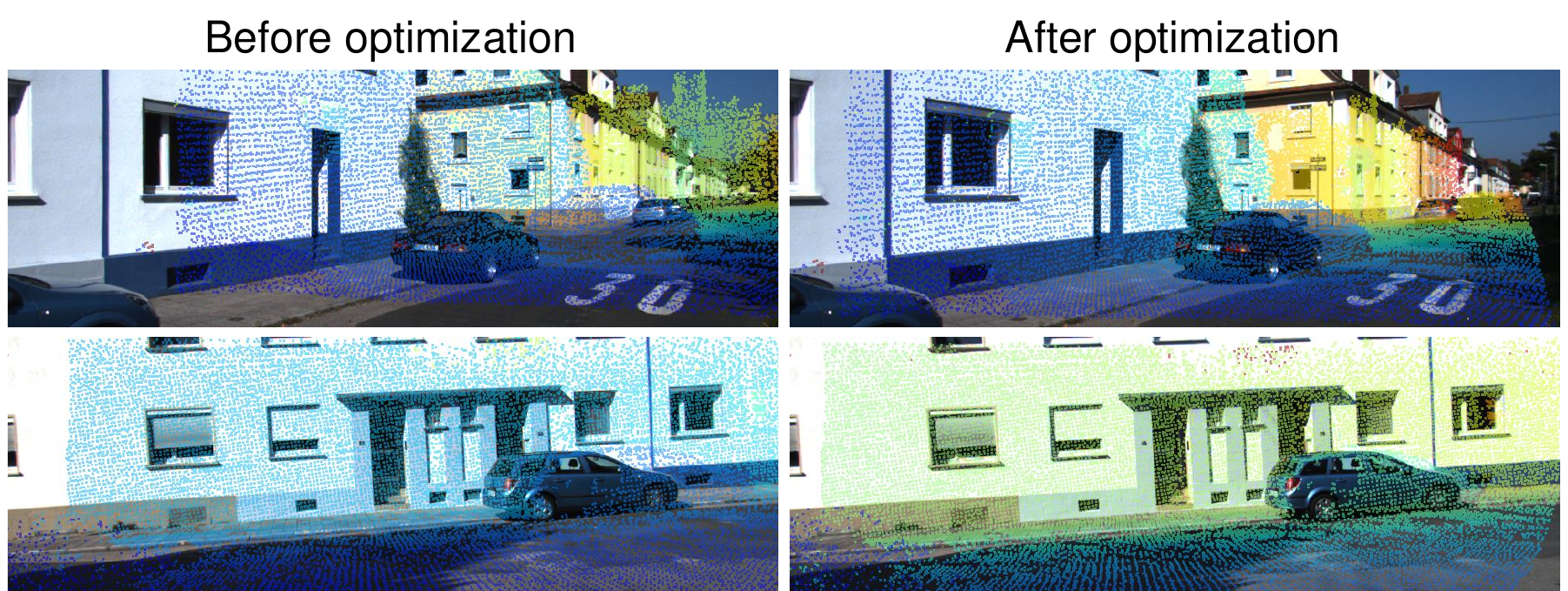}
    \caption{LiDAR projection overlaid on the next frame before and after optimization.} 
    \vspace{-3mm}
    \label{fig:optimization}
\end{figure}

\begin{table*}[htbp]
    \caption{Performance comparison on KITTI dataset.}
    \centering
    \resizebox{\linewidth}{!}{
    \begin{tabular}{c|cccc|cccc|cccc|cccc}
    \hline
    \multirow{3}{*}{Method}  & \multicolumn{4}{c|}{KITTI\_00 (1861m)} & \multicolumn{4}{c}{KITTI\_10 (459m)} & \multicolumn{4}{c|}{Arg\_2c07 (44m)} & \multicolumn{4}{c}{Arg\_2595 (54m)}\\
    & \multicolumn{2}{c|}{Transl.[m] $\downarrow$} & \multicolumn{2}{c|}{Rot.[$^\circ$] $\downarrow$} & \multicolumn{2}{c|}{Transl.[m] $\downarrow$} & \multicolumn{2}{c|}{Rot.[$^\circ$] $\downarrow$} & \multicolumn{2}{c|}{Transl.[m] $\downarrow$} & \multicolumn{2}{c|}{Rot.[$^\circ$] $\downarrow$} & \multicolumn{2}{c|}{Transl.[m] $\downarrow$} & \multicolumn{2}{c}{Rot.[$^\circ$] $\downarrow$}\\
    & Mean & Std & Mean & Std & Mean & Std & Mean & Std & Mean & Std & Mean & Std  & Mean & Std & Mean & Std\\
    \hline
    VINS-Fusion\cite{qin2018vins} & 16.70 & 9.40 & 4.99 & 2.51 & 2.34 & 1.13 & 1.75 & \bf{0.74} & 3.09 & 2.24 & 22.87 & 6.77 & 19.02 & 11.04 & 176.67 & 1.11\\
    I2D-VO & 0.24 & 0.26 & 0.54 & 0.94 & 0.61 & 0.91 & 2.74 & 15.61 & 0.43 & 0.23 & \bf{0.45} & 0.42 & 0.34 & 0.22 & 4.86 & \bf{0.18}\\
    Ours & \bf{0.13} & \bf{0.17} & \bf{0.49} & \bf{0.55} & \bf{0.45} & \bf{0.63} & \bf{1.04} & 1.24 & \bf{0.17} & \bf{0.15} & 0.58 & \bf{0.34} & \bf{0.14} & \bf{0.11} & \bf{2.96} & 0.20\\	
    \hline
    \end{tabular}
    }
    \label{tab:3}
\end{table*} 

\begin{figure*}
	\centering
	\begin{minipage}[b]{0.49\linewidth}
		\centering
		\includegraphics[width = 1.0\columnwidth]{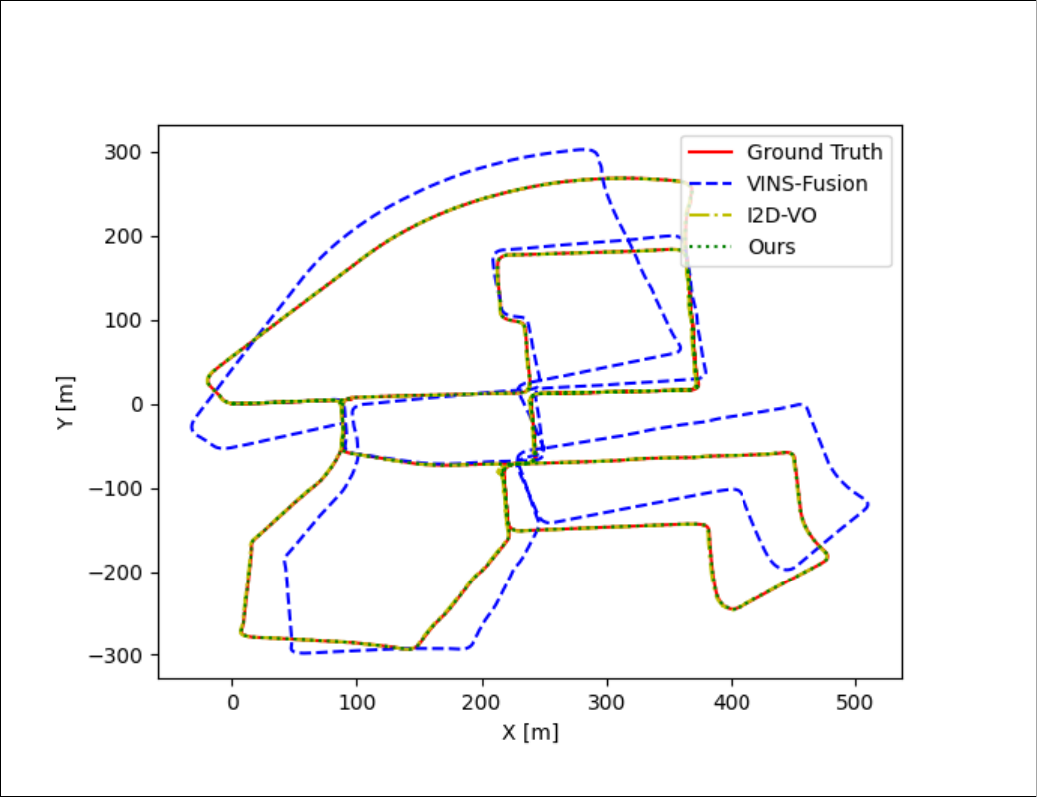} 
		\subcaption{KITTI\_00}
	\end{minipage}
	\begin{minipage}[b]{0.49\linewidth}
		\centering
		\includegraphics[width = 1.0\columnwidth]{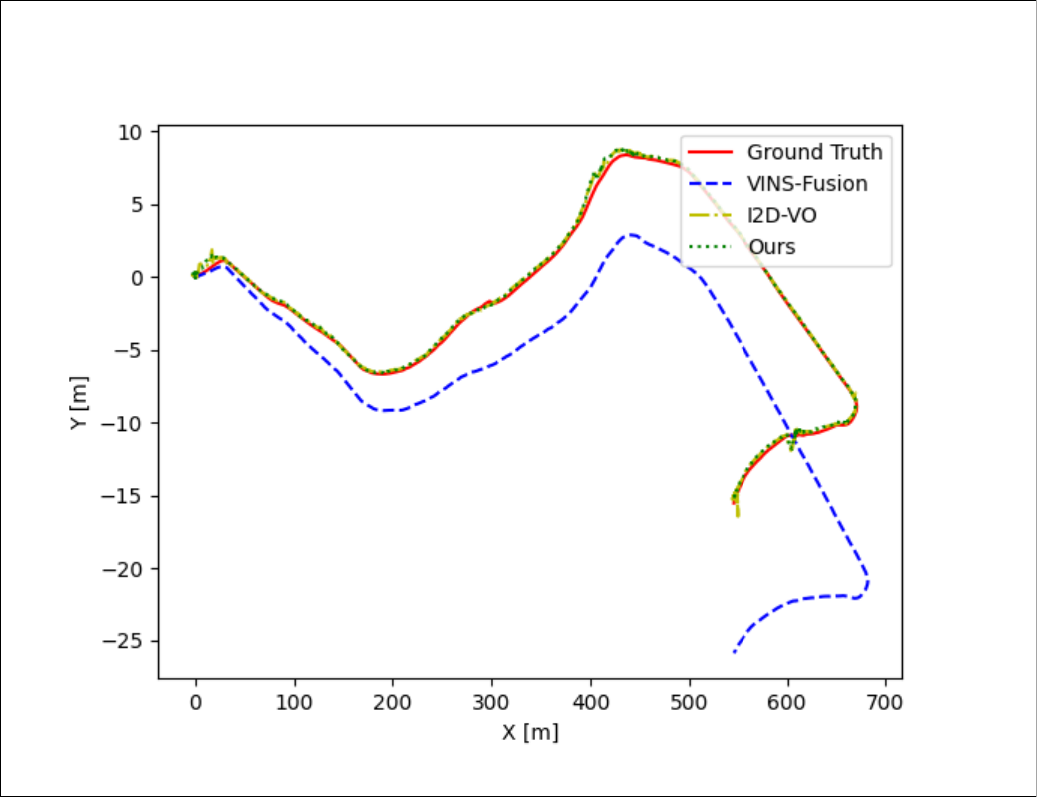} 
		\subcaption{KITTI\_10}
	\end{minipage}
	\begin{minipage}[b]{0.49\linewidth}
		\centering
		\includegraphics[width = 1.0\columnwidth]{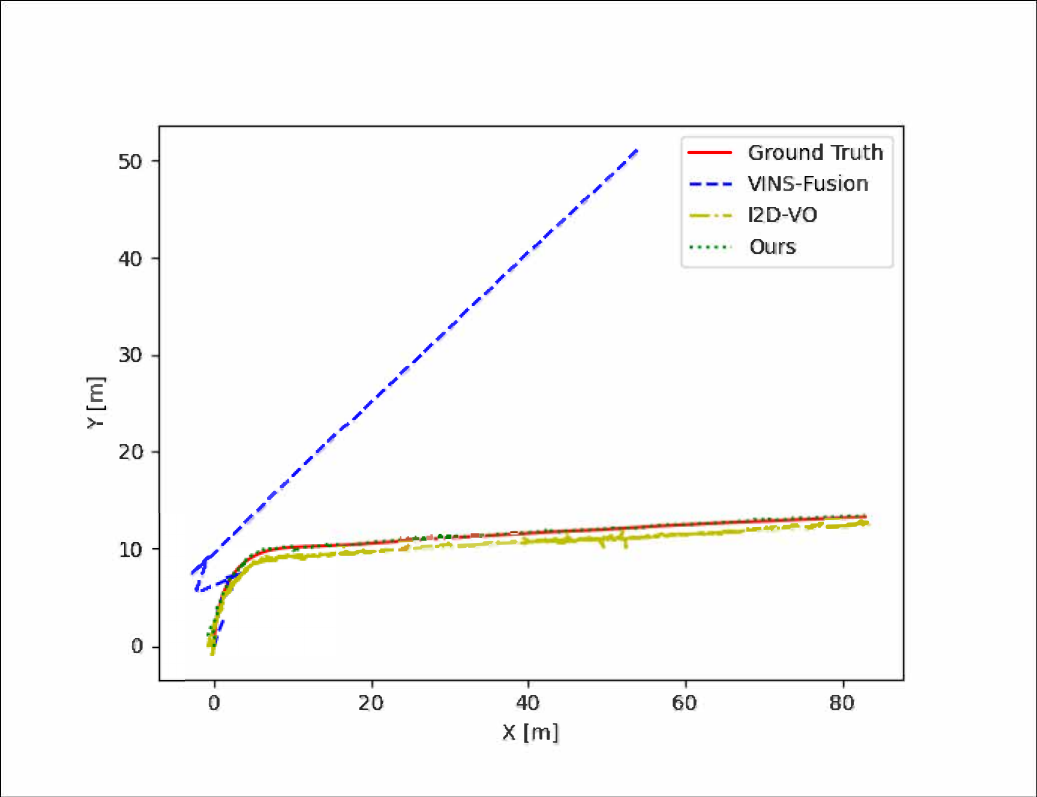} 
		\subcaption{Argoverse\_2c07}
		\label{fig:4-c}
	\end{minipage}
	\begin{minipage}[b]{0.49\linewidth}
		\centering
		\includegraphics[width = 1.0\columnwidth]{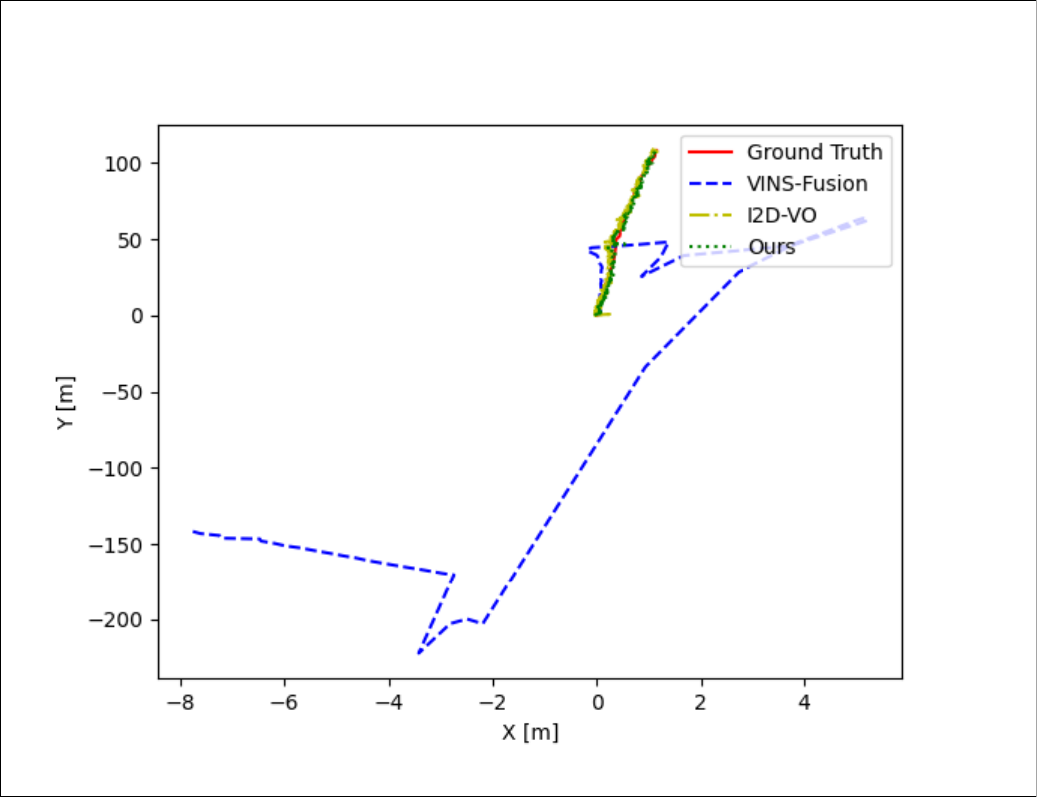} 
		\subcaption{Argoverse\_2595}
		\label{fig:4-d}
	\end{minipage}
	% \caption{Estimated trajectories for four sampled LiDAR maps.}
        \caption{Estimated trajectories of three tested methods on four sampled LiDAR maps.}
	\label{fig:trajectory}
\end{figure*}

According to Table \ref{tab:2}(a) and (b), CMRNet efficiently tracks the entire LiDAR map without any interruption, albeit with certain limitations in its localization accuracy. Conversely, I2D-Loc boasts superior localization precision, yet it falls short of completing the entire map. This difference arises due to the heavy dependence of I2D-Loc on accurate 2D-3D correspondences, while CMRNet directly regresses camera poses.
Additionally, as shown in Table \ref{tab:2}(c), the traditional visual odometry (VO) algorithm can track the entire map seamlessly. However, as shown in Fig. 1, the drift of VO is serious.
% Additionally, as shown in Table \ref{tab:2}(c), the traditional visual odometry (VO) algorithm can track the entire map seamlessly, yet it is compromised by a lower localization accuracy due to the inherent scale ambiguity.
By integrating I2D-Loc with the visual odometry algorithm in a loosely-coupled manner, our proposed framework, I2D-VO, accomplishes an enhancement in localization accuracy and successfully overcomes previous shortcomings. In Table \ref{tab:2}(e) and (f),  our proposed 2D-3D pose tracking framework, which incorporates multi-view constraints, achieves minimal translation and rotation errors upon map completion. Furthermore, as demonstrated in Fig. \ref{fig:optimization}, the accuracy of 2D-3D correspondences noticeably improves the following optimization in various challenging scenarios. In these cases, a single image-to-depth flow estimation network will fail to predict reliable 2D-3D correspondences due to homogeneous depth projections or extremely dark RGB images.
To highlight the outstanding performance of our proposed 2D-3D pose tracking methodology, we conducted a comprehensive comparison with the state-of-the-art visual-only pose tracking system, VINS-Fusion\cite{qin2018vins}. The evaluation was conducted on multiple sequences, including sequences 00 and 10 from the KITTI dataset, as well as sequences 2c07- and 2595- from the Argoverse dataset. The estimated trajectories of these sampled sequences are depicted in Fig. \ref{fig:trajectory}, demonstrating the excellent alignment of our proposed method with the ground truth trajectories.

Quantitative analysis, presented in Table \ref{tab:3}, provides clear evidence of the superiority of our proposed 2D-3D pose tracking method, which incorporates multi-view constraints. Across all four LiDAR maps, our method consistently outperforms the other tested methodologies, delivering remarkable performance.
% To underline the superior performance of our proposed 2D-3D pose tracking methodology, we have conducted a comparison with the state-of-the-art visual-only pose tracking system, VINS-Fusion\cite{qin2018vins}. The comparison is performed over multiple sequences, specifically sequences 00 and 10 from the KITTI dataset and sequences 2c07- and 2595- from Argoverse. We plot the estimated trajectories of these samples in Fig. \ref{fig:trajectory}. It can be observed that the trajectories of our proposed method align well with the ground truth.
% Table \ref{tab:3} gives the quantitative comparison and clearly illustrates that our proposed 2D-3D pose tracking method, which utilizes multi-view constraints, consistently delivers superior performance over the other tested methodologies across all four LiDAR maps.

% Finally, to demonstrate the SOTA performance of the proposed method, we conduct experiments on three LiDAR maps generated from KITTI sequences 00, 05, and 10. Among them, sequences 00 and 10 are only used for testing the generalization of the proposed method. The results are shown in Table \ref{tab:3}. We compare the proposed method with the SOTA stereo algorithm VINS-Fusion\cite{qin2018vins} on the KITTI odometry benchmark, as well as the devised loosely coupling pose tracking algorithm. The experimental results demonstrate that the proposed multi-view constraint based pose tracking algorithm outperforms the other methods on all three LiDAR maps.

\begin{figure}[htbp]
    \centering
    \includegraphics[width=0.95\linewidth]{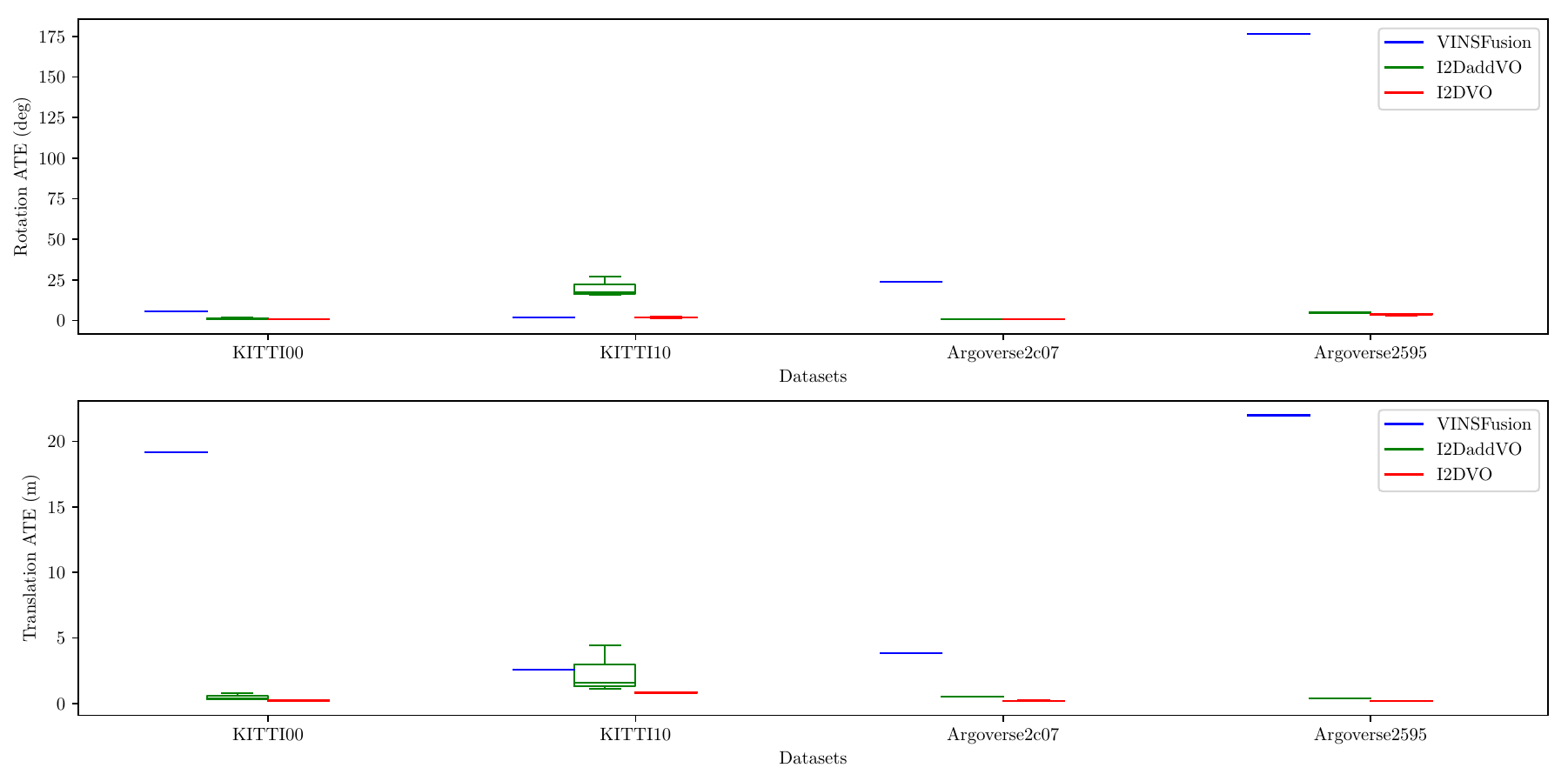}
    \caption{Comparison of ATE between VINS-Fusion and our proposed methods.} 
    \vspace{-3mm}
    \label{fig:ATE}
\end{figure}

\begin{figure}[htbp]
    \centering
    \includegraphics[width=0.95\linewidth]{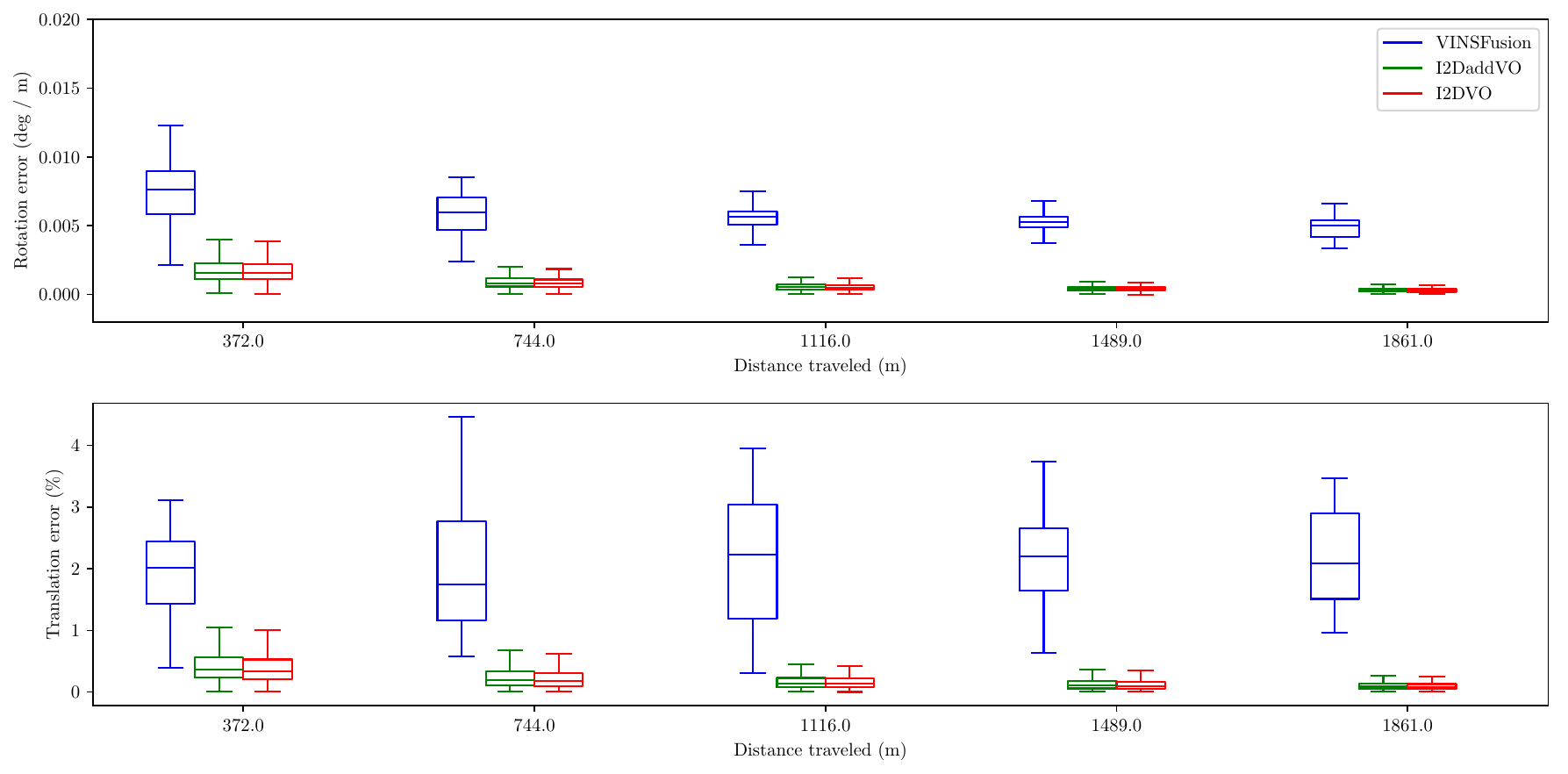}
    \caption{Comparison of RPE between VINS-Fusion and our proposed methods on the KITTI 00 sequence.} 
    \vspace{-3mm}
    \label{fig:RPE}
\end{figure}

% In addition, we also compare the proposed method with VINS-Fusion using the metrics ATE and RPE. The quantitative comparison results are shown in Fig. \ref{fig:ATE} and Fig. \ref{fig:RPE}. They demonstrate that the proposed method not only has higher localization accuracy but also has less drift during pose tracking than both the devised I2D-VO and the SOTA visual-only method VINS-Fusion.
Additionally, we have also conducted a comparison between our proposed method and VINS-Fusion, utilizing Absolute Trajectory Error (ATE) and Relative Pose Error (RPE) as the key performance metrics. The quantitative outcomes are presented in Fig. \ref{fig:ATE} and Fig. \ref{fig:RPE}. These figures emphatically demonstrate that our proposed method consistently achieves superior localization accuracy. Moreover, it exhibits reduced drift during pose tracking compared to both the carefully engineered I2D-VO and the current visual-only VINS-Fusion.
% Furthermore, we conduct a comparison between the proposed method and VINS-Fusion using the metrics ATE and RPE. The quantitative results of this comparison are depicted in Fig. \ref{fig:ATE} and Fig. \ref{fig:RPE}. The results clearly illustrate that our proposed method achieves higher localization accuracy and exhibits less drift during pose tracking compared to both the devised I2D-VO and the state-of-the-art visual-only method VINS-Fusion.

% \begin{figure}[htbp]
%     \centering
%     \includegraphics[width=1.0\linewidth]{figures/all_translation_ate.pdf}
%     \caption{Comparison of ATE between VINS-Fusion and our proposed methods.} 
%     \vspace{-3mm}
%     \label{fig:ATE}
% \end{figure}

% \begin{figure}[htbp]
%     \centering
%     \includegraphics[width=1.0\linewidth]{figures/KITTI_00_trans_rot_error.pdf}
%     \caption{Comparison of RPE between VINS-Fusion and our proposed methods on the KITTI 00 sequence.} 
%     \vspace{-3mm}
%     \label{fig:RPE}
% \end{figure}

\subsection{Discussion}
The iterative update module in I2D-Loc\cite{CHEN2022209} improves the matching accuracy by solving the large displacement problem. However, when the initial pose is already accurate enough, the motion tends to be small. This hypothesis holds in our pose tracking pipeline. Consequently, we conduct additional experiments to validate the impact of reducing the number of iterative updates. The results of these experiments are presented in Table \ref{tab:4}.  

\begin{table}[htbp]
    \caption{Performance comparison of the proposed method with different iterative numbers.}
    \centering
    \resizebox{\linewidth}{!}{
    \begin{tabular}{c|cc|cc|c}
    \hline
    \multirow{2}{*}{Num} &
    \multicolumn{2}{c|}{Translation Error} & 
    \multicolumn{2}{c|}{Rotation Error} & 
    \multirow{2}{*}{Time[ms]}\\ 
    & Mean.[cm] $\downarrow$ & Std.[cm] $\downarrow$ & Mean.[$^\circ$] $\downarrow$ & Std.[$^\circ$] $\downarrow$ &\\
    \hline
    24 & 14 & 16 & 0.49 & 0.53 & 588\\
    20 & 14 & 17 & 0.50 & 0.56 & 562\\
    16 & 14 & 17 & 0.51 & 0.57 & 540\\
    12 & 15 & 17 & 0.51 & 0.56 & 520\\
    \hline
    \end{tabular}
    }
    \label{tab:4}
\end{table}

The experimental results demonstrate that reducing the number of iterative updates can speed up the localization process without causing a significant decrease in accuracy. However, despite the improved efficiency achieved by reducing the number of iterations, our pose tracking method currently operates at a top speed of 3-4 frames per second, which is mainly due to inefficient point cloud cutting and projection.

\section{Conclusions and Future Work}
\label{sec:conclusion}
% In this work, we propose a novel 2D-3D pose tracking framework that tightly couples 2D-3D correspondences of consecutive frames using 2D-2D matching. Specifically, we devise a cross-modal consistency-based loss function to facilitate network supervision under multi-view constraints. Additionally, we define an energy function for simultaneously optimizing the poses of adjacent camera frames during the pose tracking process. Furthermore, we present a pose tracking framework that loosely couples 2D-3D and 2D-2D correspondences for comparative analysis. Through extensive experiments, we demonstrate that our proposed method significantly enhances the smoothness and accuracy of the pose tracking process.
In this study, we propose a novel 2D-3D pose tracking framework that tightly integrates 2D-3D correspondences through 2D-2D matching. Our approach incorporates a cross-modal consistency-based loss function to enable effective network supervision under multi-view constraints. Additionally, we introduce a non-linear least square problem for joint optimization of adjacent camera frame poses during the pose tracking process. Furthermore, we present a comparative analysis by incorporating a pose tracking framework that loosely couples 2D-3D and 2D-2D correspondences. Extensive experiments demonstrate that our proposed method significantly improves the smoothness and accuracy of the pose tracking. In the future, we will further extend our method to various scenarios and improve the efficiency. 

\bibliographystyle{IEEEtran.bst}
\bibliography{references}

\newpage

 

% \end{IEEEbiography}



\vfill

\end{document}